\documentclass[runningheads]{llncs}
\usepackage{graphicx}

\usepackage{tikz}
\usepackage{comment}
\usepackage{amsmath,amssymb} 
\usepackage{color}
\usepackage{algorithmic}
\usepackage{algorithm}

\usepackage{tabularx,booktabs}
\usepackage{multirow}

\begin{document}
\pagestyle{headings}
\mainmatter
\def\ECCVSubNumber{3}  

\title{Hierarchical HMM \\ for Eye Movement Classification} 

\titlerunning{Hierarchical HMM for Eye Movement Classification}
%
\author{Ye Zhu\inst{1} \and
Yan Yan\inst{1} \and
Oleg Komogortsev\inst{1}}
\authorrunning{Y. Zhu et al.}
%
\institute{Texas State University, San Marcos, USA \\
\email{\{ye.zhu, tom\_yan, ok\}@txstate.edu}
}
\maketitle

\begin{abstract}
In this work, we tackle the problem of ternary eye movement classification, which aims to separate fixations, saccades and smooth pursuits from the raw eye positional data. The efficient classification of these different types of eye movements helps to better analyze and utilize the eye tracking data.
Different from the existing methods that detect eye movement by several pre-defined threshold values, we propose a hierarchical Hidden Markov Model (HMM) statistical algorithm for detecting fixations, saccades and smooth pursuits. The proposed algorithm leverages different features from the recorded raw eye tracking data with a hierarchical classification strategy, separating one type of eye movement each time. Experimental results demonstrate the effectiveness and robustness of the proposed method by achieving competitive or better performance compared to the state-of-the-art methods.
\keywords{Hidden Markov Model, eye movement, fixation, smooth pursuit, saccade, classification}
\end{abstract}

\section{Introduction}

Eye tracking technology, which aims to measure the location where a person is looking at,
has been widely applied in various research and application fields including the human-computer interaction~\cite{poole2006eye,bruneau2002eyes}, AR/VR~\cite{duchowski2001binocular,hickson2019eyemotion}, behavioral psychology~\cite{granka2004eye,boraston2007application} and usability studies~\cite{schiessl2003eye,ehmke2007identifying} in recent years. With the arising attention and interests from researchers, eye tracking is becoming a potential and promising driver for future immersive technologies.

One fundamental and significant research topic in eye tracking is to identify different types of eye movements.
There are several primary types of eye movements: fixations correspond to the situation where the visual gaze is maintained on a single location, saccades are fast movements of the eyes that rapidly change the point of fixation, and smooth pursuits are defined as slower tracking movements of the eyes designed to keep a moving stimulus on the fovea~\cite{purves2009neuroscience}. Ternary eye movement classification~\cite{komogortsev2013automated}, which seeks to classify three primary types of eye movement, \textit{i.e.}, fixations, saccades and smooth pursuits, is essential to the above applications.

In this work, we tackle the problem of ternary eye movement classification from a probabilistic perspective by adopting the Hidden Markov Model in a hierarchical way. The hierarchical structure makes it possible to consider several different data features in different stages of classification.
The usage of the Viterbi~\cite{forney1973viterbi} and Baum-Welch algorithms~\cite{baum1970maximization} allows us to avoid the inconvenience of selecting thresholds and to improve the robustness of the classification method. Experiments show that our proposed hierarchical HMM is able to achieve competitive performance compared to the state-of-the-art methods.

\section{Related Work}
\label{sec:relatedwork}
One of the most common and intuitive methods to separate fixations from saccades is threshold based algorithms, such as
Velocity Threshold Identification (I-VT)~\cite{bahill1981variability} and Dispersion Threshold Identification (I-DT)~\cite{salvucci2000identifying}. The former method assumes the velocity of saccades should be larger than the velocity of fixations, while the latter one relies on the difference of duration and positional dispersion between fixations and saccades. However, the above single threshold-based algorithms are unable to accurately separate smooth pursuits from fixations due to a variety of artifacts usually present in the captured eye movement signal. 

The existing threshold-based state-of-the-art methods for ternary eye movement classification mainly combine several different single threshold-based algorithms. Velocity Velocity Threshold Identification (I-VVT)~\cite{komogortsev2013automated} is a basic algorithm that adopts two velocity thresholds, the data points with higher velocity than the larger velocity threshold are classified as saccades, the points with a lower velocity than the smaller threshold are classified as fixations, while the remaining points are considered as smooth pursuits. Velocity Dispersion Threshold Identification (I-VDT)~\cite{komogortsev2013automated} is another algorithm that combines I-VT and I-DT~\cite{komogortsev2013automated}. I-VDT firstly filters out saccades by I-VT, I-DT is then further used to separate fixations from smooth pursuits. Velocity Movement Pattern Identification (I-VMP)~\cite{komogortsev2010standardization} uses I-VT to identify saccades, and then employs movement direction information to separate fixations from smooth pursuits. All the methods mentioned above rely on empirically selected thresholds to provide a meaningful classification on a targeted dataset.

In addition to the above threshold-based algorithms, video-based methods to detect gazes have also been exploited in~\cite{ebisawa1998improved,li2013learning}. Dewhurst \textit{et al.}~\cite{dewhurst2012depends} proposes to use geometric vectors to detect eye movements. Nystrom \textit{et al.}~\cite{nystrom2010adaptive} aims to classify fixations, saccades and glissades using an adaptive velocity-based algorithm. Santini \textit{et al.}~\cite{santini2016bayesian} propose Bayesian method (I-BDT) to identify fixations, saccades and smooth pursuits.
Another branch of more recent work adopts machine learning techniques to tackle the problem. Identification using Random Forest machine learning technique (IRF) is used to classify fixations, saccades and post-saccadic oscillations in~\cite{zemblys2018using}. Zembly \textit{et al.}~\cite{zemblys2019gazenet} further propose a gaze-Net to realize end-to-end eye-movement event detection with deep neural networks. Startsev \textit{et al.}~\cite{startsev20191d}
tackles the problem of ternary eye movement classification with a 1D-CNN with BLSTM.

Although the machine learning methods, especially deep learning, have been widely applied in multiple research fields including the eye movement classification, one major drawback of these techniques is that the training process relies on a large amount of data. In addition, the neural networks usually require re-training when applied in a different task or dataset.
Our focus in this work is to improve the robustness and performance of the threshold-based state-of-the-art methods, even competing with the recent machine learning based state-of-the-art performance.

\section{Methodology}
\label{sec:method}

We present the proposed hierarchical HMM method in this section and provide the corresponding pseudo-code.

\begin{algorithm}[t]
\caption{Hierarchical HMM for Ternary Eye Movement Classification}
\label{algo1}
    \begin{algorithmic}[1]
    \REQUIRE Eye positional data sequence. 
    \STATE $\varepsilon_n$ number of epochs for n-th HMM, initial start probability vector $\pi$, initial transition probability matrix $\textbf{A}$ and initial emission probability matrix $\textbf{B}$. Note that initial parameters for HMM will be optimized and updated by Baum-Welch algorithm.
    \STATE $\textbf{Step 0: Pre-processing}$
    \STATE Compute the position, velocity, acceleration feature sequences for input eye tracking data
    \STATE Select appropriate features for classification.
    \STATE $\textbf{Step 1: Rough classification}$
    \STATE Initialize the parameters of the first HMM for selected feature
    \STATE $e_1 \leftarrow 0$
    \FOR{$e_1 < \varepsilon_1$}
        \STATE Viterbi algorithm
        \STATE Baum-Welch algorithm
        \STATE Filter saccades
        \ENDFOR
    \STATE $\textbf{Step 2: Refined classification}$
    \STATE Initialize the parameters of the second HMM for selected feature, define a threshold value $\mathcal{T}$ of the first feature for fine-tuning
    \STATE $e_2 \leftarrow 0$
    \FOR{$e_2 < \varepsilon_2$}
        \STATE Viterbi algorithm
        \STATE Baum-Welch algorithm
        \STATE Classify fixations and smooth pursuits
        \STATE Fine-tune the classification results by $\mathcal{T}$
        \ENDFOR
    \STATE $\textbf{Step 3: Merge function}$
    \STATE Merge classified points into complete fixations, saccades and smooth pursuits
    \RETURN List of classification results
    \end{algorithmic}
\end{algorithm}

Hidden Markov Model is a statistical model for time series based on the Markov process with hidden states. The principle of HMM is to determine the hidden state with a maximum probability according to the observable sequence. A traditional HMM takes three sets of parameters as input, which are the start probability vector, transition probability matrix and emission probability matrix. The ternary eye movement classification can be formulated as a first-order three-state HMM problem, whose hidden states are fixations, saccades and smooth pursuits.

HMM relies on the distinguishable probability distributions of features to correctly define different hidden states, otherwise, the maximum probability of each hidden state may be incorrect if several hidden states have similar probabilities given a certain observation sequence. In the case of eye movement classification, the probability distributions for different movement types are usually represented by continuous Gaussian distributions~\cite{salvucci1998tracing}. The main challenge of ternary eye movement classification using the existing methods lies in the bias of the features for smooth pursuits. While fixations and saccades have very different positional dispersion and velocity features, smooth pursuits are rather ambiguous in terms of dispersion and velocity since the steady state of a smooth pursuit often contains corrective saccades, making the eye tracking data very noisy. To this end, we propose to leverage different features and introduce a hierarchical strategy to tackle the problem. The core idea of our hierarchical HMM is to perform HMM classification for multiple times using different features.

Firstly, we start by analyzing different features of the raw eye positional data as the pre-processing step, \textit{e.g.}, positions, velocities, accelerations, whose objective is to determine the features that can be used to separate three eye movement types in latter steps.
After selecting appropriate features from pre-processing, we then perform a first-stage rough classification on the eye tracking data to separate saccades from the fixations and smooth pursuits using the HMM. We refer the first-stage as rough classification due to the reason that the classification results will be fine-tuned in the latter stage. The second-stage classification adopts a different feature to separate fixations and smooth pursuits by another HMM, using the first feature as a fine-tuning criterion at the same time. The final step of our hierarchical HMM is to merge the classified points into complete eye movements with temporal duration criterion.

The pseudo-code of the proposed method is presented in Algo.~\ref{algo1}. Compared with the existing threshold-based state-of-the-art methods, our method has several advantages: 1) the usage of HMM avoids the tedious work to select threshold values as for previous algorithms and improves the robustness of the proposed classification method; 2) a hierarchical strategy with fine-tuning technique further improves the classification performance. In the meanwhile, the proposed method does not rely on a large amount of data for training and leverages several different features in a coherent way (\textit{e.g.}, use the feature from the first stage as a fine-tuning criterion in the second stage). Experimental results prove that our proposed hierarchical HMM method is simple, straightforward yet effective.

\section{Experiments}
\label{sec:exp}

In this section, we present the experimental results obtained with the proposed method. The comparisons with other state-of-the-art methods demonstrate the robustness and effectiveness of the proposed hierarchical HMM. An ablation study is also included to show the contributions of the hierarchical structure.

\subsection{Dataset and Evaluation Metrics}
We use the eye tracking dataset recorded from the previous research work~\cite{komogortsev2013automated} for experiments. The data are recorded by the EyeLink 100 eyetracker at 1000 Hz on a 21-in monitor and contain 11 subjects, containing the human annotations as either clean or noisy data. 

Seven different behavior scores are used as quantitative evaluation metrics for our experiments. Saccade quantitative score (SQnS), fixation quantitative score (FQnS) and smooth pursuit quantitative score (PQnS) are used to measure the amount of saccades, fixation and smooth pursuit behavior in response to a stimulus, respectively~\cite{komogortsev2010standardization}. Fixation qualitative score (FQlS), smooth pursuit qualitative score for positional accuracy (PQlS\_P) and for velocity accuracy (PQlS\_V) are to compare the proximity of the detected smooth pursuit signal with the signal presented in the stimuli~\cite{komogortsev2013automated}. Misclassified fixation score (MisFix) of the smooth pursuit is defined as the ration between misclassified smooth pursuit points and the total number of fixation points in the stimuli. The ideal number of MisFix should consider the practical latency situation where smooth pursuit continues when the stimulus changes from smooth pursuit to fixation.

\subsection{Implementations}

\begin{table}[t]
\begin{center}
\caption{Comparison of behavior scores obtained by different methods for the subjects that are manually evaluated as "Medium", "Good" and "Bad", respectively. Our proposed method achieves competitive performance close to the human evaluation (\textit{i.e.}, Manual).}
\scalebox{0.75}{
\label{tab:1}
\begin{tabular}{|c|c|c|c|c|c|c|c|c|c|}
\hline
Behavior scores & Ideal & Manual~\cite{komogortsev2013automated} & Ours & I-VVT~\cite{komogortsev2013automated} & I-VDT~\cite{komogortsev2013automated} & I-VMP~\cite{komogortsev2010standardization} &
IRF~\cite{zemblys2018using} &
I-BDT~\cite{santini2016bayesian}&  1DCNN~\cite{startsev20191d} \\
\hline
SQnS & 100\% & 84\% & 83\% & 86\% & 82\% & 78\%  & 86\% & 84\% & 88\% \\ \hline
FQnS & 84\% &  63\% & 63\% & 16\% & 79\% & 66\%  & 78\% & 68\% & 59\% \\ \hline
PQnS & 52\% & 47\% & 48\% & 38\% & 55\% &  61\% & 43\% & 51\% & 40\%         \\ \hline
MixFix & 7.1\% & 13\% & 8.9\% & 42\% & 6\% &  22\% & 12\% & 15\% & 20\% \\ \hline
FQlS & $0^{\circ}$ & $0.46 ^{\circ}$ & $0.4^{\circ}$ & $0.5^{\circ}$ & $0.5^{\circ}$ & $0.5 ^{\circ}$ & $0.5^{\circ}$ & $0.5^{\circ}$ & $0.4^{\circ}$ \\ \hline
PQlS\_P  & $0^{\circ}$ & $3.07^{\circ}$ & $3.2^{\circ}$ & $3.2^{\circ}$ & $3.2^{\circ}$ &  $3.4^{\circ}$   & $3.1^{\circ}$ & $3.2^{\circ}$ & $3.2^{\circ}$ \\ \hline
PQlS\_V & $0^{\circ}/s$ & $39^{\circ}/s$ & $30^{\circ}/s$ & $16^{\circ}/s$ & $47^{\circ}/s$ & $40^{\circ}/s$ & $38^{\circ}/s$ & $33^{\circ}/s$ & $40 ^{\circ}/s$ \\ \hline
\end{tabular}
}
\end{center}

\begin{center}
\scalebox{0.75}{
\begin{tabular}{|c|c|c|c|c|c|c|c|c|c|}
\hline
Behavior scores & Ideal & Manual~\cite{komogortsev2013automated} & Ours & I-VVT~\cite{komogortsev2013automated} & I-VDT~\cite{komogortsev2013automated} & I-VMP~\cite{komogortsev2010standardization} &
IRF~\cite{zemblys2018using} &
I-BDT~\cite{santini2016bayesian} & 
1DCNN~\cite{startsev20191d}\\
\hline
SQnS & 100\% & 96\% & 91\% & 96\% & 90\% & 90\% & 88\% & 92\% & 90\% \\ \hline
FQnS & 84\% &  71\% & 74\% & 30\% & 82\% & 69\% & 82\% & 80\% & 78\%\\ \hline
PQnS & 52\% & 39\% & 44\% & 39\% & 30\% &  56\%  & 58\% & 42\% & 38\%   \\ \hline
MixFix & 7.1\% & 6\% & 5.5\% & 50\% & 4.4\% &  21\% & 12\% & 8\% & 16\% \\ \hline
FQlS & $0^{\circ}$ & $0.44^{\circ}$ & $0.3^{\circ}$ & $0.4^{\circ}$ & $0.4^{\circ}$ & $0.4^{\circ}$ & $0.4^{\circ}$ & $0.4^{\circ}$ & $0.4^{\circ}$ \\ \hline
PQlS\_P  & $0^{\circ}$ & $3.15^{\circ}$ & $2.9^{\circ}$ & $3.6^{\circ}$ & $3.2^{\circ}$ &  $3.7^{\circ}$  & $3.7^{\circ}$ & $3.3^{\circ}$ & $3.4^{\circ}$ \\ \hline
PQlS\_V & $0^{\circ}/s$ & $23^{\circ}/s$ & $25^{\circ}/s$ & $16^{\circ}/s$ & $44^{\circ}/s$ & $40^{\circ}/s$ & $38^{\circ}/s$ & $37^{\circ}/s$ & $32^{\circ}/s$\\ \hline
\end{tabular}
}
\end{center}

\begin{center}
\scalebox{0.75}{
\begin{tabular}{|c|c|c|c|c|c|c|c|c|c|}
\hline
Behavior scores & Ideal & Manual~\cite{komogortsev2013automated} & Ours & I-VVT~\cite{komogortsev2013automated} & I-VDT~\cite{komogortsev2013automated} & I-VMP~\cite{komogortsev2013automated} &
IRF~\cite{zemblys2018using} &
I-BDT~\cite{santini2016bayesian}& 
1DCNN~\cite{startsev20191d}\\
\hline
SQnS & 100\% & 89\% & 77\% & 85\% & 78\% & 76\% & 75\% & 82\% &  85\% \\ \hline
FQnS & 84\% &  42\% & 49\% & 17\% & 61\% & 51\% & 43\%& 45\% & 50\% \\ \hline
PQnS & 52\% & 40\% & 37\% & 33\% & 31\% &  47\% & 44\% & 38\% & 33\%     \\ \hline
MixFix & 7.1\% & 33\% & 11\% & 33\% & 22\% &  36\% & 18\% & 12\% & 20\%\\ \hline
FQlS & $0^{\circ}$ & $0.58^{\circ}$ & $0.5^{\circ}$ & $0.6^{\circ}$ & $0.6^{\circ}$ & $0.6^{\circ}$ & $0.6^{\circ}$ & $0.6^{\circ}$ & $0.6^{\circ}$ \\ \hline
PQlS\_P  & $0^{\circ}$ & $2.58^{\circ}$ & $2.6^{\circ}$ & $3.2^{\circ}$ & $3.5^{\circ}$ &  $3.6^{\circ}$  & $3.3^{\circ}$ & $2.4^{\circ}$ & $3.5^{\circ}$ \\ \hline
PQlS\_V & $0^{\circ}/s$ & $30^{\circ}/s$ & $32^{\circ}/s$ & $16^{\circ}/s$ & $81^{\circ}/s$ & $52^{\circ}/s$ & $46^{\circ}/s$ & $31^{\circ}/s$ & $44^{\circ}/s$\\ \hline
\end{tabular}}
\end{center}
\end{table}

We compare our proposed hierarchical HMM method with six baselines: I-VVT~\cite{komogortsev2013automated}, I-VDT~\cite{komogortsev2013automated}, I-VMP~\cite{komogortsev2010standardization}, IRF~\cite{zemblys2018using}, I-BDT~\cite{santini2016bayesian} and 1DCNN~\cite{startsev20191d}. Among six baselines, I-VVT, I-VDT, and I-VMP are threshold-based methods, all the threshold values are optimized for different subject recordings in our experiments for comparison.
For the IRF, we use the same selected features (\textit{i.e.}, velocity and position) as for our proposed method. For the I-BDT, we follow the parameters chosen in~\cite{santini2016bayesian}. For the 1DCNN, we take the pre-trained model and fine-tune it on the recording data (except those used for testing) from our targeted dataset.

For our proposed hierarchical HMM method, we use the parameters reported in~\cite{salvucci1998tracing} for the initialization in HMM. We iterate the Viterbi and Baum-Welch algorithms for 3 times to learn and update the parameters for HMM. For the final Merge function, the merge time interval threshold we use is 75ms, the merge distance is 0.5$^{\circ}$.

\subsection{Experimental Results}
\noindent \textbf{Results on Behavior Scores.} We compare the behavior scores obtained via the proposed hierarchical HMM and other methods on the subjects that are manually evaluated as "Medium", "Good" and "Bad" in the eye tracking data quality~\cite{komogortsev2013automated}. The ideal scores and manually classification results are also reported for comparison, the calculation of the ideal behavior scores considers the actual physiological reactions of humans when performing different types of eye movements.

\begin{figure}[t]
    \centering
    \includegraphics[width=0.98\textwidth]{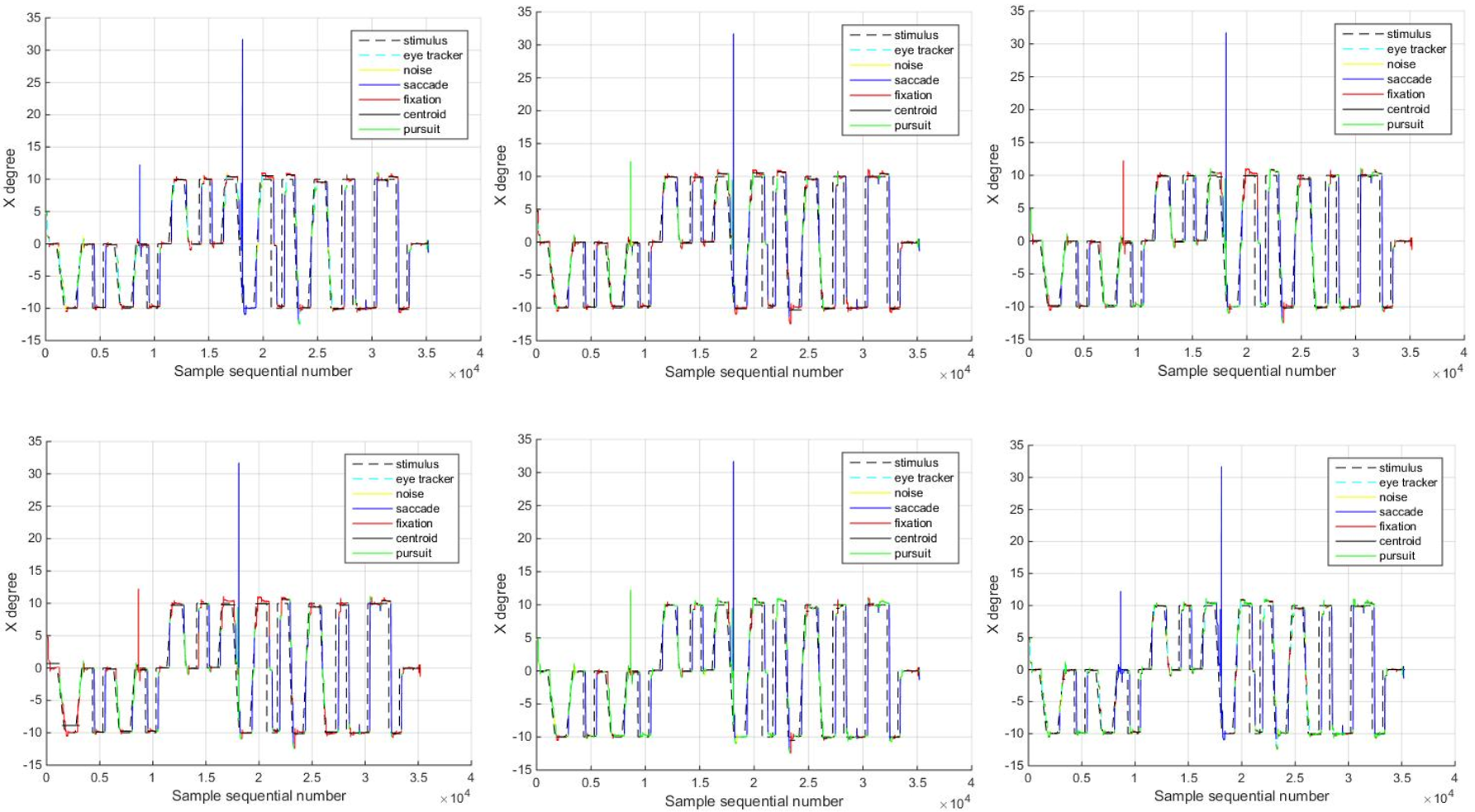}
    \caption{Qualitative classification results on the "good quality" subject obtained by the proposed hierarchical HMM, I-VDT, I-VMP, IRF, I-BDT and 1D CNN, respectively. Best viewed in color.}
    \label{fig:fig2}
\end{figure}

The quantitative experimental results are presented in Table~\ref{tab:1}. We observe that our proposed hierarchical HMM method achieves promising performance compared to the other baseline methods and is much closer to human performance (\textit{i.e.}, manual). The I-VVT algorithm has the worst performance as expected due to the reason that it only considers the velocity and separates the eye movement by two simple threshold values. The I-VDT and I-VMP are able to separate fixations, saccades and smooth pursuits, and I-VDT has better performance compared to I-VMP, achieving a MixFix score closer to the ideal one.
Other machine learning based methods IRF, I-BDT, and 1DCNN achieve good performance in general, however, we observe that 1DCNN has relatively lower behavior scores in detecting the smooth pursuit. One possible reason for its under-performance could be the limited amount of smooth pursuit data for fine-tuning. Since the original model was trained in a different dataset and fine-tuned in our experiments, the transfer learning process may be less effective compared to retraining the entire network model, which also reveals one of the limitations about those "data-driven" methods.
Our proposed hierarchical HMM outperforms the other considered methods, especially the threshold-based methods, in most of the behavior scores and it is worth noting that the proposed method has behavior scores close to the manually evaluated results.

We also present the qualitative classification results in Fig.~\ref{fig:fig2}. Due to the reason that the I-VVT basically fails in this ternary eye movement classification task, we do not include the classification results of I-VVT in the figure. The qualitative results are consistent with the previous quantitative results. Notably, our proposed hierarchical HMM method succeeds in detecting an unexpected saccade while most of the other methods detect it either as smooth pursuit or fixation.

\noindent \textbf{Feature Analysis.} In our experiments, we propose a pre-processing step to analyze different features from raw data.
We analyze velocity, position and acceleration features, and select velocity and position as two features used for the hierarchical HMM method. Velocity feature is used in the first HMM to filter saccades and also as the fine-tuning criterion in the second stage HMM, and the position features are used as the main feature to separate fixations and smooth pursuits in the second stage classification.

\begin{figure}[t]
    \centering
    \includegraphics[width=0.98\textwidth]{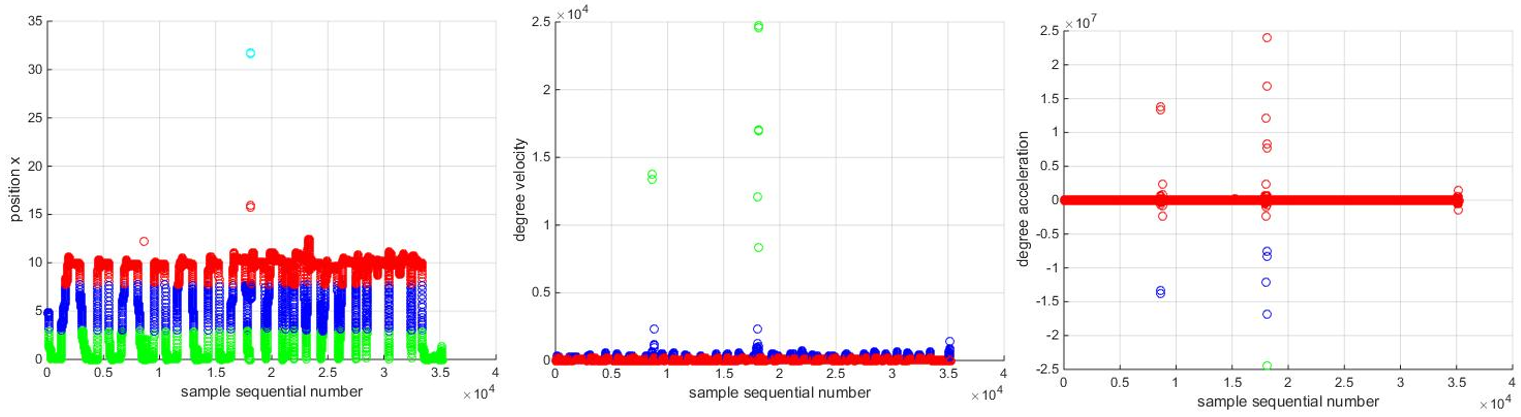}
    \caption{Visualization of clustering results for position, velocity and acceleration features, respectively, on the data recording from the subject manually evaluated as "good"~\cite{komogortsev2013automated}. Different colors represent different clusters automatically detected by K-means. Best viewed in color.}
    \label{fig:fig1}
\end{figure}

We present the visualization results of the data features on the subject that is manually evaluated as "good" in~\cite{komogortsev2013automated} using the K-Means cluster algorithm. The main objective of the pre-processing is to select appropriate features for further classifications using HMM. In the meanwhile, it also provides some insights for the reasons why the classic threshold-based methods may fail to achieve good performance in the ternary eye movement classification task. As shown in Fig.~\ref{fig:fig1}, the actual eye tracking data usually contain a lot of noises. Take the velocity feature as an example, there exist some noisy data points with extremely large velocities and the difference of velocity among fixations, saccades and smooth pursuits are not very clear. Therefore, the statistical methods with predefined threshold values are more likely to fail in this case. 

\noindent \textbf{Ablation Study Analysis.} We conduct an additional ablation experiment to demonstrate the necessity and effectiveness of the proposed hierarchical structure. Since the fixations, saccades and smooth pursuits in the ternary eye movement classification can be considered as three hidden states of an HMM, we use the velocity feature as the criterion and adopt a single-stage HMM with the three-hidden state to do the same task, in other words, we remove the second stage classification from the proposed method.  

\begin{table}[t]
\begin{center}
\caption{Comparison of behavior scores for the subject manually evaluated as "good" in the ablation study. Our proposed hierarchical strategy especially contributes to the classification of smooth pursuits.}
\scalebox{1.0}{
\label{tab:2}
\begin{tabular}{|c|c|c|c|c|}
\hline
Behavior scores & Ideal & Manual~\cite{komogortsev2013automated} & Ours & 3-state HMM \\
\hline
SQnS & 100\% & 96\% & 91\% & 91\%  \\ \hline
FQnS & 84\% &  71\% & 74\% & 54\%  \\ \hline
PQnS & 52\% & 39\% & 44\% & 25\%            \\ \hline
MixFix & 7.1\% & 6\% & 5.5\% & 28\%      \\ \hline
FQlS & $0^{\circ}$ & $0.44 ^{\circ}$ & $0.3^{\circ}$ & $0.4^{\circ}$   \\ \hline
PQlS\_P  & $0^{\circ}$ & $3.15^{\circ}$ & $2.9^{\circ}$ & $3.4^{\circ}$     \\ \hline
PQlS\_V & $0^{\circ}/s$ & $23^{\circ}/s$ & $25^{\circ}/s$ & $31^{\circ}/s$  \\ \hline
\end{tabular}}
\end{center}
\end{table}

\begin{figure}[t]
    \centering
    \includegraphics[width=0.9\textwidth]{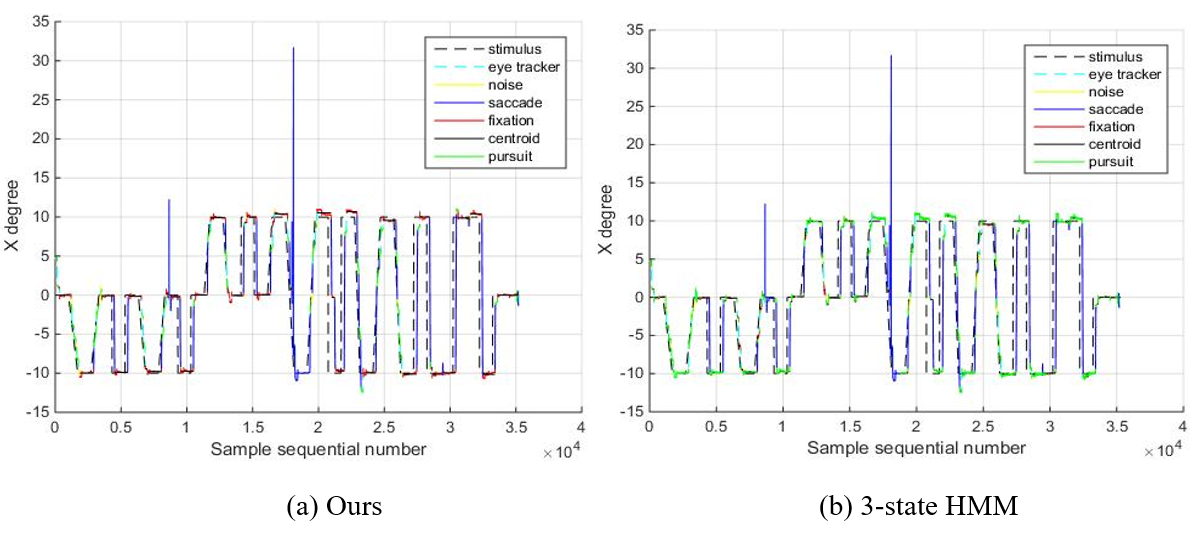}
    \caption{Qualitative results for the ablation study. Our proposed hierarchical strategy helps with the classification of smooth pursuits, which is consistent with the quantitative results.}
    \label{fig:fig3}
\end{figure}

The quantitative and qualitative results are shown in Table~\ref{tab:2} and Fig.~\ref{fig:fig3}, respectively. Both experimental results prove that the hierarchical structure contributes to better classification results. The qualitative figure shows that the hierarchical structure helps especially with the separation between fixations and smooth pursuits, which further validates the fact that the velocity feature is biased for fixations and smooth pursuits. Therefore, the analysis of different features and the usage of position features in the second stage of the proposed method are necessary and beneficial.

\section{Conclusions}
\label{sec:conclusion}

In this paper, we propose a hierarchical HMM algorithm to tackle the problem of ternary eye movement classification from the raw eye positional data. Different features from the data are considered to realize the multi-stage hierarchical classification. Experiments on multiple data records demonstrate the effectiveness and robustness of the proposed method. Possible future directions of this work could be incorporating the pre-processing step into the proposed method in an automated way, and making efforts to classify more categories of eye movements with a deeper hierarchical structure.  


\clearpage
%
%
\bibliographystyle{splncs04}
\bibliography{eccv2020submission}
\end{document}